\documentclass{SCIS2022}

\def\mathbi#1{\textbf{\em #1}}
\usepackage[figuresright]{rotating}
\usepackage{algorithm}
\usepackage{algorithmic}
\usepackage{multirow}
\usepackage{amssymb}
\usepackage{subfig}
\usepackage{newfloat}
\usepackage{listings}
\usepackage{booktabs}
\usepackage{color}
\usepackage{array}
\usepackage{graphicx}
\usepackage{array}

\begin{document}

\ArticleType{RESEARCH PAPER}
\Year{2022}
\Month{}
\Vol{}
\No{}
\DOI{}
\ArtNo{}
\ReceiveDate{}
\ReviseDate{}
\AcceptDate{}
\OnlineDate{}

\title{Span-based joint entity and relation extraction augmented with sequence tagging mechanism}{}

\author[$\dag$]{Bin JI}{}
\author[$\dag$]{Shasha LI}{}
\author[*]{Hao XU}{xuhao@nudt.edu.cn}
\author[]{Jie YU}{}
\author[]{Jun MA}{}
\author[*]{Huijun LIU}{liuhuijun@nudt.edu.cn}
\author[]{Jing YANG}{}

\AuthorMark{Ji B}

\AuthorCitation{Ji B, Li S S, Xu H,  et al.}

\contributions{Bin Ji and Shasha Li have the same contribution to this work.}

\address[]{College of Computer, National University of Defense Technology, Changsha {\rm 410073}, China}

\abstract{
Span-based joint extraction simultaneously conducts named entity recognition (NER) and relation extraction (RE) in text span form. 
However, since previous span-based models rely on span-level classifications, they cannot benefit from token-level label information, which has been proven advantageous for the task.
In this paper, we propose a \textbf{S}equence \textbf{T}agging augmented \textbf{S}pan-based \textbf{N}etwork (STSN), a span-based joint model that can make use of token-level label information. 
In STSN, we construct a core neural architecture by deep stacking multiple attention layers, each of which consists of three basic attention units.
On the one hand, the core architecture enables our model to learn token-level label information via the sequence tagging mechanism and then uses the information in the span-based joint extraction; on the other hand, it establishes a bi-directional information interaction between NER and RE.
Experimental results on three benchmark datasets show that STSN consistently outperforms the strongest baselines in terms of F1, creating new state-of-the-art results.
}

\keywords{joint extraction, named entity recognition, relation extraction, span, sequence tagging mechanism}

\maketitle

\section{Introduction}
\begin{figure}[t]
\centering
\includegraphics[width=0.8\textwidth]{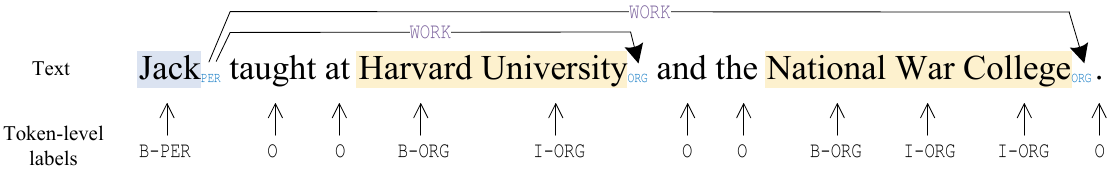} 
\caption{A span-based joint extraction example, which contains three gold entities and two gold relations. Tokens in shade are span examples, \texttt{PER} and \texttt{ORG} are entity types, \texttt{WORK} is a relation type. We also label the text with token-level labels via the sequence tagging mechanism, such as \texttt{B-PER}, \texttt{B-ORG} etc.}
\label{figure1}
\end{figure}

The joint entity and relation extraction task extracts both entities and semantic relations between entities from raw texts. 
It acts as a stepping stone for a variety of downstream NLP tasks \cite{li_ji}, such as question answering. 
According to the classification methods, we {divide} the existing models for the task into two categories: sequence tagging-based models \cite{mi_ba,ka_ca,wei,lin} and span-based models \cite{luan18,di_al,eb_ul,ji,chendanqi}. The former is based on the sequence tagging mechanism and performs token-level classifications. The latter is based on the span-based paradigm and performs span-level classifications. 
Since the sequence tagging mechanism and the span-based paradigm are considered to be distinct methodologies, existing joint extraction models permit the use of just one of them.
Specifically, the span-based paradigm consists of three typical steps: it first splits raw texts into text spans (a.k.a. candidate entities), such as the ``Jack" and ``Harvard University" in Figure \ref{figure1}; it then constructs ordered span pairs (a.k.a. candidate relation tuples), such as the $<$``Jack", ``Harvard University"$>$ and $<$``Harvard University", ``Jack"$>$; and finally, it jointly classifies spans and span pairs. 
For example, it classifies the ``Jack" and ``Harvard University" into \texttt{PER} and \texttt{ORG}, respectively. And it classifies the $<$``Jack", ``Harvard University"$>$ and $<$``Harvard University", ``Jack"$>$ into \texttt{WORK} and \texttt{NoneType}, respectively.\footnote{The span-based paradigm assigns the \texttt{NoneType} to spans that are not entities, as well as span pairs that do not hold relations.}

The majority of span-based models \cite{di_al,eb_ul,chendanqi} use Pre-trained Language Models (PLMs) as their encoders directly, which relies on the encoding ability of PLMs heavily, resulting in insufficient span semantic representations and poor model performance. 
To alleviate this problem, some span-based models \cite{luan,wadden} make attempts to incorporate other related NLP tasks into this task, such as event detection and coreference resolution. By using carefully designed neural architectures, these models enable span semantic representation to incorporate information shared from the added tasks.
However, these additional tasks require extra data annotations such as event annotations, which are inaccessible in most datasets for the task, such as SciERC \cite{luan18}, DocRED \cite{docred}, TACRED \cite{tacred}, NYT \cite{nyt}, WebNLG \cite{webnlg}, SemEval \cite{semeval}, CoNLL04 \cite{conll}, and ADE \cite{ade} etc.

Previous sequence tagging-based joint models \cite{mi_ba,wei,be,zhao} demonstrate that token-level labels convey critical information, which can be used to compensate for span-level semantic representations. 
For example, if a span-based model is aware that the “Jack” is a person entity (labeled with the \texttt{PER} label) and the {“Harvard University”} is an organization entity (labeled with the \texttt{ORG} label) beforehand, it may readily infer that they have a \texttt{WORK} relation. 
Unfortunately, as far as we know, existing span-based models neglect this critical information due to their inability to produce token-level labels. 
Additionally, existing sequence tagging-based models establish a unidirectional information flow from NER to RE by using the token-level label information in the relation classification, hence enhancing information sharing. Due to the lack of token-level labels, previous span-based models are unable to build such an information flow, let alone a more effective bi-directional information interaction.

In this paper, we explore using the token-level label information in the span-based joint extraction, aiming to improve the performance of the span-based joint extraction.
To this end, we propose a \textbf{S}equence \textbf{T}agging augmented \textbf{S}pan-based \textbf{N}etwork (STSN) where the core module is a carefully designed neural architecture, which is achieved by deep stacking multiple attention layers. 
Specifically, the core architecture first learns three types of semantic representations: label representations for classifying token-level labels, and token representations for span-based NER and RE, respectively; it then establishes information interactions among the three learned representations. 
As a result, the two types of token representations can fully incorporate label information. Thus, span representations constructed with the above token representations are also enriched with label information.
Additionally, the core architecture enables our model to build an effective bi-directional information interaction between NER and RE.

For the above purposes, each attention layer of the core architecture consists of three basic attention units:
(1) Entity\&Relation to Label Attention (\textbf{E\&R-L-A}) enables label representations to attend to the two types of token representations. The reason for doing this is two-fold: one is that E\&R-L-A enables label representations to incorporate task-specific information effectively; the other is that E\&R-L-A is essential to construct the bi-directional information interaction between NER and RE.
(2) Label to Entity Attention (\textbf{L-E-A}) enables token representations for NER to attend to label representations with the goal of enriching the token representations with label information.  
(3) Label to Relation Attention (\textbf{L-R-A}) enables token representations for RE to attend to label representations with the goal of enriching the token representations with label information.
{In addition, we} establish the bi-directional information interaction by taking the label representation as a medium, enabling the two types of token representations to attend to each other. We have validated the effectiveness of the bi-directional information interaction in Section \ref{4.4.2}.
Moreover, to enable STSN to use token-level label information of overlapping entities, we extend the BIO tagging scheme and discuss more details in Section \ref{4.1.2}.

In STSN, aiming to train token-level label information in a supervised way, we add a sequence tagging-based NER decoder to the span-based model. And we use entities and relations extracted by the span-based model to evaluate the model performance. 
Experimental results on ACE05, CoNLL04, and ADE demonstrate that STSN consistently outperforms the strongest baselines in terms of F1, creating new state-of-the-art performance.\footnote{For reproducibility, our code for this paper will be publicly available at \url{https://github.com/jibin/STSN}.}

In sum, we summarize the contributions as follows: (1)
We propose an effective method to augment the span-based joint entity and relation extraction model with the sequence tagging mechanism. (2) We carefully design the deep-stacked attention layers, enabling the span-based model to use token-level label information and establish a bi-directional information interaction between NER and RE. (3) Experimental results on three datasets demonstrate that STSN creates new state-of-the-art results.

\section{Related work}

\subsection{Span-based joint extraction} 

Models for span-based joint entity and relation extraction have been widely studied. 
Luan et al. \cite{luan18} propose almost the first published span-based model, which is drawn from two models for coreference resolution \cite{lee1} and semantic role labeling \cite{he}, respectively. 
With the advent of Pre-trained Language Models (PLMs), span-based models directly take PLMs as their encoders, such as Dixit and Al-Onaizan \cite{di_al} propose a span-based model which takes ELMo \cite{elmo} as the encoder; Eberts and Ulges \cite{eb_ul} propose SpERT, which takes BERT \cite{bert} as the encoder; Zhong and Chen \cite{chendanqi} propose PURE which takes ALBERT \cite{albert} as the encoder. 
However, these models rely heavily on the encoding ability of PLMs, leading to insufficient span semantic representations and finally resulting in poor model performance. 
Some models \cite{luan,wadden} make attempts to alleviate this issue by adding additional NLP tasks to the task, such as coreference resolution or event detection. 
These models enable span semantic representations to incorporate information derived from the added tasks through complicated neural architectures. 
However, the added tasks need extra data annotations (such as event annotations are required in joint entity-relation extraction datasets), which are unavailable in most cases. 
Compared to these models, our model enriches span semantic representations with token-level label information without additional data annotations. 

\subsection{Token-level label} 

Numerous work has demonstrated that token-level label information benefits the joint extraction task a lot. 
For example,  the models reported in the literature \cite{mi_ba,ka_ca,wei,be} train fixed-size semantic representations for token-level labels and use them in relation classification by concatenating them to relation semantic representations, delivering promising performance gains. 
However, Zhao et al. \cite{zhao} demonstrate that the above shallow semantic concatenation cannot make full use of the label information. 
Therefore, they carefully design a deep neural architecture to capture fine-grained token-label interactions and deep infuse token-level label information into token semantic representations, delivering more promising performance gains. 
Unfortunately, previous span-based joint extraction models cannot benefit from the token-level label information since they completely give up the sequence tagging mechanism. 
In contrast, we propose a sequence tagging augmented span-based joint extraction model, which generates token-level label information via the sequence tagging mechanism and further infuses the information into token semantic representations via deep infusion.

\section{Approach}

\begin{figure*}[t]
\centering
\includegraphics[width=1\textwidth]{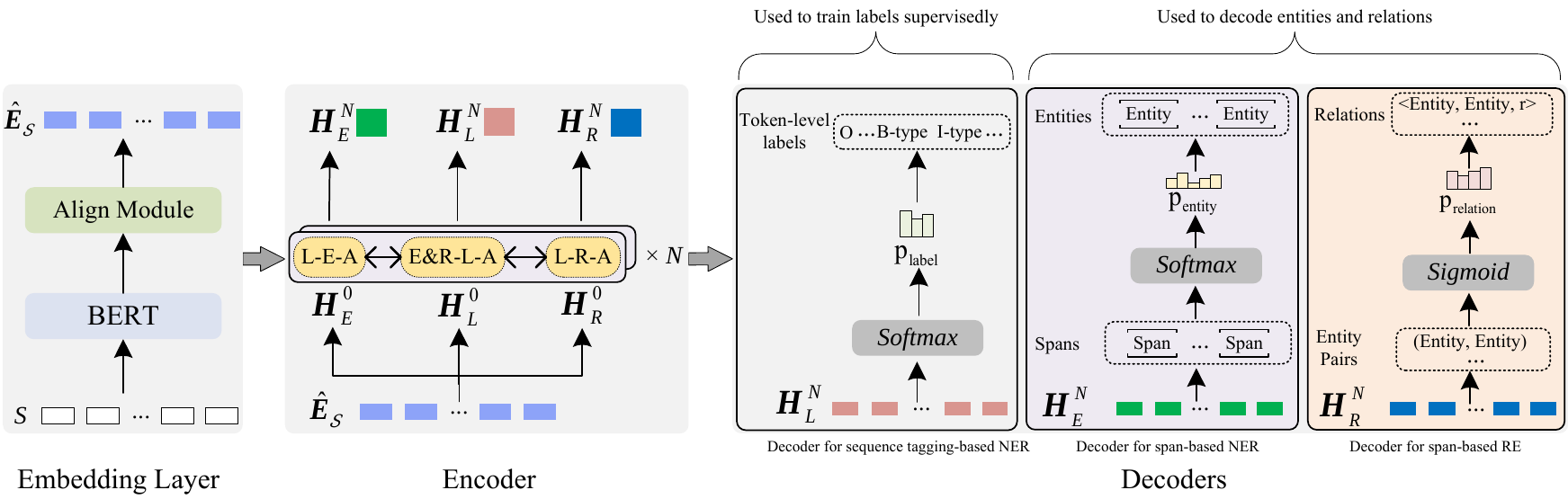} 
\caption{The illustration of STSN, which consists of a BERT-based embedding layer, an encoder and three separate linear decoders. We {solely} use the decoder for sequence tagging-based NER to train token-level label semantics ($\mathbi{H}_L$) in a supervised way. And entities and relations decoded by the span-based NER and RE decoders are used to evaluate model performance.}
\label{figure2}
\end{figure*}

In this section, we will describe the Sequence Tagging augmented Span-based Network (STSN) in detail. 
As Figure \ref{figure2} shows, STSN consists of three components: a BERT-based embedding layer, an encoder composed of deep-stacked attention layers, and three separate linear decoders for sequence tagging-based NER, span-based NER and span-based RE, respectively. 

\subsection{Embedding layer}

In STSN, we use BERT \cite{bert} as the default embedding generator.
For a given text $\scriptsize{\mathcal{T} = (t_1,  t_2, t_3,..., t_n)}$ where $t_i$ denotes the $i$-$th$ token, BERT first tokenizes it with the WordPiece vocabulary \cite{wordpiece} to obtain an input sequence. 
For each element of the sequence, its representation is the element-wise addition of WordPiece embedding, positional embedding, and segment embedding. 
Then a list of input embeddings $\mathbi{H} \in \mathbb{R}^{len \ast hid}$ are obtained, where \textit{len} is the sequence length and \textit{hid} is the size of hidden units. 
A series of pre-trained Transformer \cite{transformer} blocks are then used to project $\mathbi{H}$ into a BERT embedding sequence (denoted as $\mathbi{E}_\mathcal{T}$):
\begin{equation}
{\mathbi{E}_\mathcal{T} = \{\mathbi{e}_1, \mathbi{e}_2, \mathbi{e}_3,..., \mathbi{e}_{len}\}}.
\end{equation}

BERT may tokenize one token into several sub-tokens to alleviate the Out-of-Vocabulary (OOV) problem, leading to that $\mathcal{T}$ cannot align with $\mathbi{E}_\mathcal{T}$, i.e., $n \neq len$.
To achieve alignment, we propose an Align Module, which applies the max-pooling function to the BERT embeddings of tokenized sub-tokens to obtain token embeddings.  
We denote the aligned embedding sequence for $\mathcal{T}$ as:
\begin{equation}
{\hat{\mathbi{E}}_\mathcal{T} = \{\hat{\mathbi{e}}_1, \hat{\mathbi{e}}_2, \hat{\mathbi{e}}_3,..., \hat{\mathbi{e}}_{n}\}},
\end{equation}

\noindent where ${\hat{\mathbi{E}}_\mathcal{T}\in \mathbb{R}^{n*d}}$ and $d$ is the BERT embedding dimension. $\hat{\mathbi{e}}_i$ denotes the BERT embedding of $t_i$.

\subsection{Encoder}

The encoder is a deep neural architecture, which is achieved by stacking multiple ($N$) attention layers in depth.

\subsubsection{Deep neural architecture}

We deep stack multiple attention layers to build the deep neural architecture, where each layer is composed of three basic attention units, as shown in Figure \ref{figure2}.

The deep neural architecture learns three types of semantic representations: label representations (denoted as $\mathbi{H}_L$) used to classify token-level labels for sequence tagging-based NER, token representations (denoted as $\mathbi{H}_E$) for span-based NER, and token representations (denoted as $\mathbi{H}_R$) for span-based RE. The three representations have the same embedding dimension $d$. Additionally, we denote the concatenation of $\mathbi{H}_E$ and $\mathbi{H}_R$ as $\mathbi{H}_C$ and convert its embedding dimension to $d$ via a Feed Forward Network (FFN):
\begin{equation}
\mathbi{H}_C = [\mathbi{H}_E; \mathbi{H}_R]\mathbi{W}_C + \mathbi{b}_C,
\end{equation}

\noindent where $\mathbi{W}_C$ $\in$ $\mathbb{R}^{2d*d}$ and $\mathbi{b}_C$ $\in$ $\mathbb{R}^d$ are trainable FFN parameters.

We formulate the first attention layer as follows:
\begin{equation}
[Layer]^1 = \left\{ \begin{array}{lrc}
\mathbi{H}_C^0 = [\mathbi{H}_E^0; \mathbi{H}_R^0]\mathbi{W}_C^0 + \mathbi{b}_C^0, \\
\mathbi{H}_L^1 = E\&R$-$L$-$A(\mathbi{H}_L^0, \mathbi{H}_C^0, \mathbi{H}_C^0),\\
\mathbi{H}_E^1 = L$-$E$-$A(\mathbi{H}_E^0, \mathbi{H}_L^1, \mathbi{H}_L^1),\\
\mathbi{H}_R^1 = L$-$R$-$A(\mathbi{H}_R^0, \mathbi{H}_L^1, \mathbi{H}_L^1), 
\end{array}\right.
\end{equation}

\noindent where $\hat{\mathbi{E}}_\mathcal{T}$ is mapped to $\mathbi{H}_L^0$, $\mathbi{H}_E^0$, and $\mathbi{H}_R^0$, respectively. $\mathbi{H}_L^1$, $\mathbi{H}_E^1$, and $\mathbi{H}_R^1$ are the outputs of the first layer.

Then $ \mathbi{H}_L^1$, $ \mathbi{H}_E^1$, and $ \mathbi{H}_R^1$ are passed to the next layer. We recursively repeat the above procedure until we obtain the outputs of the $N$-$th$ layer, namely $ \mathbi{H}_L^N$, $ \mathbi{H}_E^N$, and $ \mathbi{H}_R^N$.
Now we assume that $ \mathbi{H}_E^N$ and $ \mathbi{H}_R^N$ have fully incorporated token-level label information. And they will be used for span-based NER and RE, respectively. $ \mathbi{H}_L^N$ will be used to classify token BIO labels for sequence tagging-based NER.

As Figure \ref{figure2} shows, we establish information interactions among the three types of representations in each attention layer. Specifically, $\mathbi{H}_E$ and $\mathbi{H}_L$ can interact with each other directly, as well as $\mathbi{H}_R$ and $\mathbi{H}_L$. Therefore by taking $\mathbi{H}_L$ as a medium, $\mathbi{H}_E$ and $\mathbi{H}_R$ can also interact with each other, which establishes a bi-directional information interaction between span-based NER and span-based RE in essence. 

\begin{figure}[t]
\centering
\includegraphics[width=0.6\textwidth]{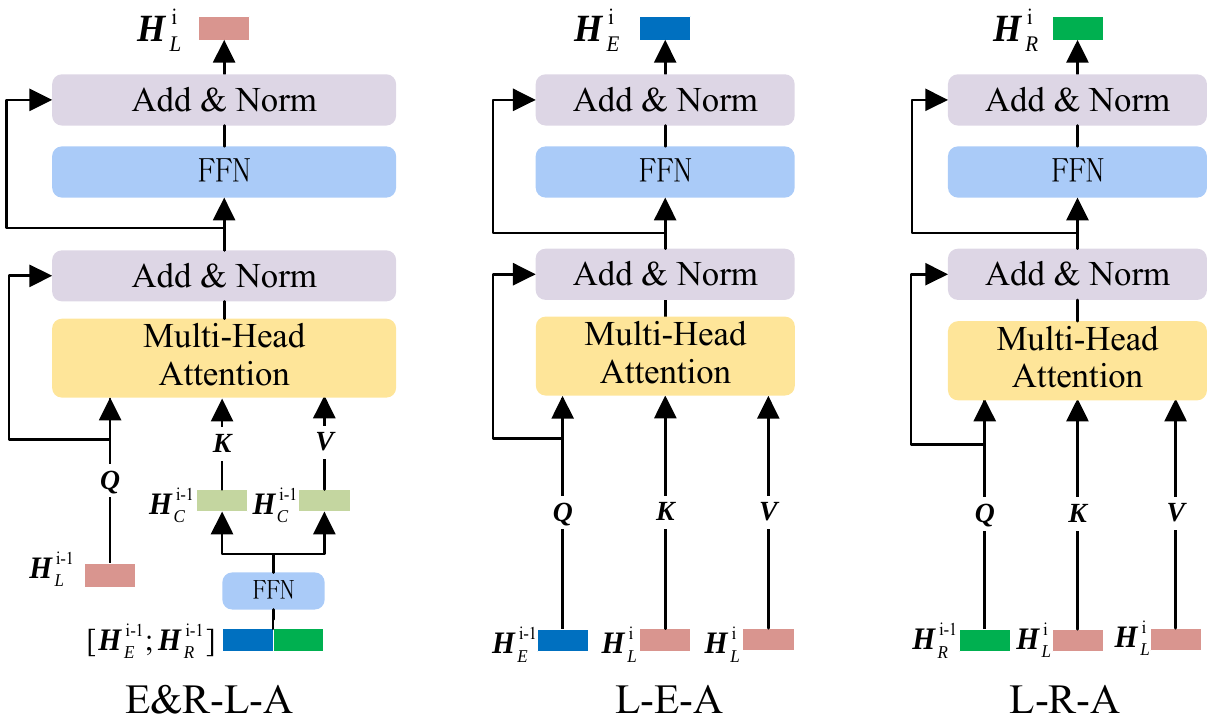} 
\caption{Neural architectures of E\&R-L-A, L-E-A, and L-R-A. The three units share a common architecture but differ in inputs.}
\label{figure3}
\end{figure}

\subsubsection{Basic attention units}

As Figure \ref{figure3} shows, the three types of basic attention units share a common neural architecture but differ in model inputs. 
 The common architecture is composed of two sub-layers: multi-head attention and position-wise FFN. A residual connection is adopted around each sub-layer, followed by layer normalization. 

Multi-head attention has been proven effective in capturing long-range dependencies by explicitly attending to all positions in various feature spaces.
It has a series of $h$ parallel heads and requires three inputs, i.e., Query $(\mathbi{Q})$, Key $(\mathbi{K})$ and Value $(\mathbi{V})$:
\begin{equation}
head^i = \textrm{softmax}(\frac{(\mathbi{Q}\mathbi{W}^i_{Q})(\mathbi{K}\mathbi{W}^i_{K})^T}{\sqrt{d/h}}(\mathbi{V}\mathbi{W}^i_{V})), 
\end{equation}
\begin{equation}
\mathbi{I} = \textrm{concat}(head^1,...,head^h)\mathbi{W}_O,
\end{equation}

\noindent  where $\{\mathbi{Q}, \mathbi{K}, \mathbi{V}\} \in \mathbb{R}^{n*d}$, $\{\mathbi{W}^i_Q, \mathbi{W}^i_K, \mathbi{W}^i_V\} \in {\mathbb{R}^{d*(d/h)}}$ and $\mathbi{W}_O \in \mathbb{R}^{d*d}$ are trainable parameters. $\mathbi{I} \in \mathbb{R}^{n*d}$ is the output.
Multi-head attention learns the pairwise relationship between $\mathbi{Q}$ and $\mathbi{K}$ and outputs weighted summation across all instances.
Then residual connection conducts element-wise addition of $\mathbi{I}$ and $\mathbi{Q}$. 

Position-wise FFN contains two linear transformations with a ReLU activation in between:
\begin{equation}
\textrm{FFN}(\mathbi{I}) = \textrm{max}(0, \mathbi{I}\mathbi{W}_1 + \mathbi{b}_1)\mathbi{W}_2 + \mathbi{b}_2,
\end{equation}

\noindent where $\{\mathbi{W}_1, \mathbi{W}_2\} \in \mathbb{R}^{d*d}$ and $\{\mathbi{b}_1, \mathbi{b}_2\} \in \mathbb{R}^d$ are trainable FFN parameters. 

Figure \ref{figure3} shows the detailed implementations of the three units. To be specific, (1) Entity\&Relation to Label Attention (E\&R-L-A) takes $\mathbi{H}_L$ as $\mathbi{Q}$, {and} $\mathbi{H}_C$ as $\mathbi{K}$ and $\mathbi{V}$, respectively. {It enables} label representations to attend to the two types of token representations, 
aiming to make label representations incorporate task-specific information well. 
(2) Label to Entity Attention (L-E-A) takes $\mathbi{H}_E$ as $\mathbi{Q}$, and $\mathbi{H}_L$ as $\mathbi{K}$ and $\mathbi{V}$, respectively. It enables token representations for span-based NER to attend to label representations, aiming to infuse label information into the token representations.
(3) Label to Relation Attention (L-R-A) takes $\mathbi{H}_R$ as $\mathbi{Q}$, and $\mathbi{H}_L$ as $\mathbi{K}$ and $\mathbi{V}$, respectively. It enables token representations for span-based RE to attend to label representations, aiming to infuse label information into the token representations.

\subsection{Decoders}

We design three separate linear decoders for sequence tagging-based NER, span-based NER and RE, respectively. 

\subsubsection{Decoder for sequence tagging-based NER} This encoder aims to train label representations in a supervised way.
 The decoder first uses an FFN to convert the {embedding} space of label representations ($d$) to the embedding space of BIO labels. It then uses the softmax function to calculate probability distributions on the BIO label space:
\begin{equation}
\hat{\mathbi{y}}_L = {\textrm{softmax}}(\mathbi{H}_L^N \mathbi{W}_L + \mathbi{b}_L),
\end{equation}

\noindent where $\mathbi{W}_L \in \mathbb{R}^{d*l}$ and $ \mathbi{b}_L \in \mathbb{R}^{l}$ are trainable FFN parameters. $l$ is the count of BIO label types.

The training objective is to minimize the following cross-entropy loss:
\begin{equation}
{\mathcal{L}_L = -\frac{1}{M_{L}} \sum \limits_{i=1}^{M_{L}}{\mathbi{y}}^i_{{L}}\log\hat{\mathbi{y}}^i_{{L}}},
\end{equation}

\noindent where ${\mathbi{y}}_{L}$ is the one-hot vector of {the} gold token BIO label. $M_L$ is the count of token-label instances. 

\subsubsection{Decoder for span-based NER}
This decoder classifies span representations to obtain entities. These entities will be used for RE and model performance evaluation.
We first add the \texttt{NoneEntity} type to the pre-defined entity types. Our model will be trained to classify spans into \texttt{NoneEntity} if they are not entities.
We formulate definition of span as:
\begin{equation}
{s = (t_i, t_{i+1}, t_{i+2}...,t_{i+j})} \quad s.t. \ \ 1 \leq i \leq i+j \leq n,
\end{equation}

\noindent where span width is restricted by a threshold $\epsilon$ and $j < \epsilon$.  
We obtain the span representation of $s$ (denoted as $\mathbi{E}_s$) by concatenating semantic representations of span head and tail tokens, and the span width embedding:
\begin{equation}
\mathbi{E}_{s} = [\mathbi{H}_{E,i}^N; \mathbi{H}_{E,i+j}^N; \mathbi{W}_{j+1}],\label{eq11}
\end{equation}

\noindent where $\mathbi{H}_{E,i}^N$ and $\mathbi{H}_{E,i+j}^N$ are the $i\mbox{-}th$ and $(i+j)\mbox{-}th$ embeddings in $ \mathbi{H}_{E}^N$. $\mathbi{W}_{j+1}$ is the fixed-size span width embedding, which is trained during model training.

$ \mathbi{E}_{s} $ first passes through an FFN then is fed into the softmax function, yielding a posterior on the space of entity types (including \texttt{NoneEntity}):
\begin{equation}
{\hat{\mathbi{y}}_{s} = softmax({\mathbi{E}}_{s}\mathbi{W}_{s} + \mathbi{b}_{s})},
\end{equation}

\noindent where $\mathbi{W}_{s}$ and $\mathbi{b}_{s}$ are trainable FFN parameters. 
The training objective is to minimize the following cross-entropy loss:
\begin{equation}
{\mathcal{L}_E = -\frac{1}{M_{E}} \sum \limits_{i=1}^{M_{E}}{\mathbi{y}}^i_s\log\hat{\mathbi{y}}^i_s},
\end{equation}

\noindent where ${\mathbi{y}}_s$ is the one-hot vector of the gold span type. ${M_{E}}$ is the number of span instances. 

We filter spans that are predicted as entities and build an entity set $\mathbb{S}_e$.

\subsubsection{Decoder for span-based RE} 
This decoder classifies relation representations to obtain relations. These relations will be used for model performance evaluation.
As relations exist between entities, only spans predicted as entities are used for the classification.
We formulate the definition of ordered entity pairs (a.k.a. candidate relation tuple) as: 
\begin{equation}
{r} = <{e}_1, {e}_2> \quad s.t. \ \ e_1, e_2 \in \mathbb{S}_e, \   e_1 \neq e_2,
\end{equation}

\noindent where $e_1$ and $e_2$ are the head and tail entities, respectively. 

We obtain relation representations (denoted as $\mathbi{E}_r$) by concatenating semantic representations of head entity, tail entity and relation context:
\begin{equation}
\mathbi{E}_{r} = [\mathbi{E}_{{e}_1}; \mathbi{E}_{{e}_2}; \mathbi{C}_{r}],
\end{equation}

\noindent where $\mathbi{E}_{{e}_1}$ and $\mathbi{E}_{{e}_2}$ are semantics of ${e}_1$ and ${e}_2$, respectively. 
We obtain them using Eq.\ref{eq11} with $\mathbi{H}_{R}^N$.
Following previous work \cite{eb_ul}, we obtain $\mathbi{C}_{r}$ by applying the max-pooling function to the embedding sequence of the relation context.

$ \mathbi{E}_{r} $ first passes through an FFN then is fed into the sigmoid function:
\begin{equation}
{\hat{\mathbi{y}}_{{r}} = \sigma({\mathbi{E}}_{r}\mathbi{W}_{r} + \mathbi{b}_{r})},
\end{equation}

\noindent where $\sigma$ is the sigmoid. $\mathbi{W}_{r}$ and $\mathbi{b}_{r}$ are trainable FFN parameters. 

Any high response in the sigmoid outputs indicates that a corresponding relation is held between $e_1$ and $e_2$. Given a confidence threshold $\alpha$, any relation with a $score \ge \alpha$ is considered activated. 

The training objective is to minimize the following binary cross-entropy loss:
\begin{equation}
{\mathcal{L}_R = -\frac{1}{M_{R}} \sum \limits_{i=1}^{M_{R}}(\mathbi{y}^i_{r}\log\hat{\mathbi{y}}^i_{r} + (1- \mathbi{y}^i_{r}) \log(1-\hat{\mathbi{y}}^i_{r}))},
\end{equation}

\noindent where ${\mathbi{y}}_{{r}}$ is the one-hot vector of the gold relation type. ${M_{R}}$ is the number of entity pair instances.

\subsubsection{Model training}
During model training, we optimize the following joint training objective:
\begin{equation}
\mathcal{L}_{joint}(\mathbi{W};\theta) = \mathcal{L}_L + \mathcal{L}_E + \mathcal{L}_R.
\end{equation}

\begin{figure}[t]
\centering
\includegraphics[width=0.45\textwidth]{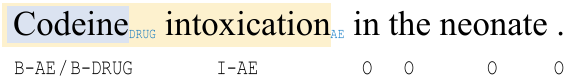} 
\caption{An example of overlapping entities which are tagged by the extended BIO tagging scheme, where ``Codeine'' is the overlapping token, \texttt{DRUG} and \texttt{AE} are entity types.}
\label{figure4}
\end{figure}

\section{Experiments}

\subsection{Experimental setup}

\subsubsection{Datasets}We evaluate STSN on ACE05 \cite{ace}, CoNLL04 \cite{conll}, and ADE \cite{ade} and use the same entity and relation types, data splits, and pre-processing following the established line of work  \cite{wa_lu}.
Moreover, for a fair comparison with previous work \cite{eb_ul}, we maintain a full version of the ADE dataset, which includes 119 instances containing overlapping entities.

\subsubsection{Extended BIO tagging scheme} \label{4.1.2}
To make STSN use token-level label information of overlapping entities, we extend the BIO tagging scheme, which cannot tag overlapping entities initially.
We begin by establishing two definitions:

\begin{itemize}
\item {D\scriptsize{EFINITION}} \normalsize{1}. $\mathbi{Two-fold overlapping entities}$. A pair of overlapping entities where the overlapping tokens are not contained in any other entities.
\item D\scriptsize{EFINITION} \normalsize{2}. $\mathbi{Preceding entity}$. An entity with a preceding head location. If two entities have the same head location, the entity with a longer length is chosen.
\end{itemize}

Figure \ref{figure4} gives a typical example: “Codeine” and “Codeine intoxication” are two-fold overlapping entities, and “Codeine intoxication” is the preceding entity.

The detailed tagging principle is that we first tag the preceding entity with the BIO tagging scheme. Then for the overlapping entity, we append its BIO labels to existing labels, separated by “/”.  
For example, “Codeine” is tagged with \texttt{B-AE/B-DRUG}.  
As all overlapping entities in the full ADE dataset are two-fold, we tag the dataset with the extended BIO tagging scheme.  For other datasets, we tag them with the BIO tagging scheme.

\subsubsection{Evaluation metrics}Following the established line of work \cite{ji,chendanqi}, we use the standard precision (P), recall (R) and F1 to evaluate the model performance. For NER, a predicted entity is considered correct if its type and boundaries (entity head for ACE05) match the ground truth. For RE, we adopt two evaluation metrics: (1) A predicted relation is considered correct if the relation type and boundaries of the two entities match the ground truth. We denote this metric as RE. (2)  A predicted relation is considered correct if both relation type and the two entities match ground truth. We denote this metric as RE+. More discussion of evaluation settings can be found in the literature \cite{wa_lu}.

\subsubsection{Implementation details}
We build STSN by deep stacking three attention layers and evaluate it with \texttt{bert-base-cased} \cite{bert} and \texttt{albert-xxlarge-v1} \cite{albert} on a single NVIDIA RTX 3090 GPU. We optimize STSN using AdamW for 100 epochs with a learning rate of 5$e$-5, a linear scheduler with a warmup ratio of 0.1 and a weight decay of 1$e$-2. We set the training batch size to 4, dimension of $\mathbi{W}_{j+1}$ to 150, $h$ of multi-head attention to 8, span width threshold $\epsilon$ to 10 and relation threshold $\alpha$ to 0.4.
Following the established line of work \cite{eb_ul,ji}, we adopt a negative sampling strategy and set the number of the negative entity and relation samples per data entry to 100, respectively.

Across all the three datasets, we use the training set to train STSN and use the test set to report model evaluation performance. For ACE05 and CoNLL04, we run STSN 20 times and report averaged results of the best 5 runs. For ADE, we adopt the 10-fold cross-validation and run each fold 20 times and report averaged results of the best 5 runs.

\renewcommand\arraystretch{0.8}
\begin{table}[htb] 
\centering
\setlength{\tabcolsep}{1.5mm}{
\begin{tabular}{lrcccccccccccc}
\toprule
\multirow{2}{*}{Model}         &\multirow{2}{*}{PLM}       &            & \multicolumn{3}{c}{NER}        &  & \multicolumn{3}{c}{RE}               &  & \multicolumn{3}{c}{RE+}  \\ \cmidrule{4-6} \cmidrule{8-10} \cmidrule{12-14} 
                                                                  &       &                     & P          & R          & F1   &  & P          & R          & F1         &  & P    & R          & F1   \\ \midrule
Li and Ji \cite{li_ji}             &-  &                                       & 85.2       & 76.9       & 80.8 &  & 68.9       & 41.9       & 52.1       &  & 65.4 & 39.8       & 49.5 \\
Katiyar \textit{et al.} \cite{ka_ca}   &-&                    & 84.0       & 81.3       & 82.6 &  & 57.9       & 54.0       & 55.9       &  & 55.5 & 51.8       & 53.6 \\
Miwa \textit{et al.} \cite{mi_ba}      &-&                    & 82.9       & 83.9       & 83.4 &  & -          & -          & -          &  & 57.2 & 54.0       & 55.6 \\
Sun \textit{et al.} \cite{sun}             &-&                 & 83.9       & 83.2       & 83.6 &  & -          & -          & -          &  & 64.9 & 55.1       & 59.6 \\
Li \textit{et al.} \cite{li}   &\texttt{BERT}   &      & 84.7       & 84.9       & 84.8 &  & -          & -          & -          &  & 64.8 & 56.2       & 60.2 \\
Dixit and Al \cite{di_al}       &\texttt{ELMo}  &                                   & 85.9       & 86.1       & 86.0 &  & 68.0       & 58.4       & 62.8       &  & -    & -          & -    \\
Shen \textit{et al.} \cite{shen}    &\texttt{BERT} &    & 87.7       & 87.5       & 87.6 &  & \textbf{-} & -          & \textbf{-} &  & 62.2 & 63.7       & 62.8 \\
Luan \textit{et al.} \cite{luan}           &- &                & -          & \textbf{-} & 88.4 &  & -          & -          & -          &  & -    & \textbf{-} & 63.2 \\
Wadden \textit{et al.} \cite{wadden}   &\texttt{BERT}&   & \textbf{-} & \textbf{-} & 88.6 &  & \textbf{-} & \textbf{-} & \textbf{-} &  & -    & \textbf{-} & 63.4 \\
Lin \textit{et al.} \cite{lin}      &\texttt{BERT}  &   & -          & \textbf{-} & 88.8 &  & -          & -          & -          &  & -    & \textbf{-} & 67.5 \\
Wang and Lu \cite{wa_lu}       &\texttt{ALBERT}&                         & \textbf{-} & \textbf{-} & 89.5 &  & -          & -          & 67.6       &  & -    & \textbf{-} & 64.3 \\
Ji \textit{et al.} \cite{ji}    &\texttt{BERT} &       & 89.3       & 89.9       & 89.6 &  & -          & \textbf{-} & -          &  & \textbf{71.2} & 60.2       & 65.2 \\
Ren \textit{et al.} \cite{renli}     &\texttt{ALBERT}&                        & -          & \textbf{-} & 89.9 &  & -          & -          & -          &  & -    & \textbf{-} & 68.0 \\
Zhong \textit{et al.} \cite{chendanqi}      &\texttt{BERT} &      & -          & \textbf{-} & 90.1 &  & -          & \textbf{-} & 67.7       &  & -    & \textbf{-} & 64.8\\
Zhong \textit{et al.} \cite{chendanqi}    &\texttt{ALBERT}&     & -          & \textbf{-} & 90.9 &  & -          & \textbf{-} & 69.4       &  & -    & \textbf{-} & 67.0 \\
 \midrule
STSN (Ours)                  & \texttt{BERT}&                                                & {90.9}           &{89.9}            & {90.4} &  &77.8            &60.7            & 68.2       &  &69.4      & 64.4           & 66.8 \\
STSN (Ours)                & \texttt{ALBERT}  &                                              &\textbf{92.7}            & \textbf{90.5}           & \textbf{91.6} &  & \textbf{80.2}           &   \textbf{64.2}         & \textbf{71.3}       &  &{69.5}      &\textbf{68.7}            & \textbf{69.1} \\ \bottomrule
\end{tabular}}
\caption{Model comparisons on ACE05 using the micro-averaged F1. Bold values denote the state-of-the-art {results}.}\label{table1}
\end{table}

\renewcommand\arraystretch{0.8}
\begin{table}[h] 

\centering
{\setlength{\tabcolsep}{1.5mm}{
\begin{tabular}{lrlccccccc}
\toprule
\multirow{2}{*}{Model}                     & \multirow{2}{*}{PLM}    &  & \multicolumn{3}{c}{NER}                                                  &                      & \multicolumn{3}{c}{RE+}                                                        \\ \cmidrule{4-6} \cmidrule{8-10} 
                                           &                         &  & P                     & R                     & F1                       &                      & P                        & R                        & F1                       \\ \midrule
Bekoulis \textit{et al.} \cite{be} $\blacktriangle$  & -                       &  & 83.4                  & 84.1                  & 83.9                     &                      & 63.8                     & 60.4                     & 62.0                     \\
Nguyen \textit{et al.} \cite{ng_ve} $\blacktriangle$   & -                       &  & -                     & -                     & 86.2                     &                      & -                        & -                        & 64.4                     \\
Eberts \textit{et al.} \cite{eb_ul} $\blacktriangle$   & \texttt{BERT}                    &  & 85.8                  & 86.8                  & 86.3                     &                      & 74.8                     & 71.5                     & 72.9                     \\
Wang and Lu \cite{wa_lu} $\blacktriangle$  & \texttt{ALBERT}                   &  & -                     & {-}            & 86.9                     &                      & -                        & {-}               & 75.4                     \\
Miwa \textit{et al.} \cite{mi_sa}   $\vartriangle$                            & -                       &  & 81.2                  & 80.2                  & 80.7                     &                      & 76.0                     & 50.9                     & 61.0                     \\
Zhang \textit{et al.} \cite{zhang} $\vartriangle$   & -                       &  & -                     & -                     & 85.6                     &                      & -                        & -                        & 67.8                     \\
Li \textit{et al.} \cite{li} $\vartriangle$      & \texttt{BERT}                    &  & 89.0                  & 86.6                  & 87.8                     &                      & 69.2                     & 68.2                     & 68.9                     \\
Eberts \textit{et al.} \cite{eb_ul}  $\vartriangle$  & \texttt{BERT}                    &  & 88.3                  & 89.6                  & 88.9                     &                      & 73.0                     & 70.0                     & 71.5                     \\
Wang and Lu \cite{wa_lu} $\vartriangle$ & \texttt{ALBERT}                  &  & {-}            & {-}            & 90.1                     &                      & -                        & {-}               & 73.6                     \\
Ji\textit{et al.} \cite{ji} $\vartriangle$       & \texttt{BERT}                    &  & 90.1                  & 90.4                  & 90.2                     &                      & 77.0                     & 71.9                     & 74.3                     \\
Shen \textit{et al.} \cite{shen} $\vartriangle$      & \texttt{BERT}                    &  & 90.3                  & {90.3}         & 90.3                     &                      & 73.0                     & {71.6}            & 72.4                     \\
Zhao \textit{et al.} \cite{zhao} $\vartriangle$       & \texttt{ELMO}                    &  & {-}            & {-}            & 90.6                     &                      & -                        & {-}               & 73.0                     \\
Cabot \textit{et al.} \cite{cabot} $\vartriangle$     & \texttt{BART}                    &  & {-} & {-} & {-}    & {} & {75.6} & {75.1} & {75.4} \\ \midrule
STSN (Ours) $\vartriangle$             & {\texttt{BERT}}   &  &90.6                       &\textbf{91.2}                       & 90.9                     &                      & 76.1                         &73.9                          & 75.0                     \\
STSN (Ours) $\vartriangle$             & {\texttt{ALBERT}} &  &\textbf{92.4}                       &                      {90.8} & \textbf{91.6}                     &                      & \textbf{76.8}                         & \textbf{77.4}                         & \textbf{77.1}                     \\
\midrule
STSN (Ours)  $\blacktriangle$             &   {\texttt{BERT}}                      &  & 88.5                      &87.9                       & 88.2                     &                      & 77.5                         & 77.1                         & 77.3                     \\
STSN (Ours) $\blacktriangle$             & {\texttt{ALBERT}}                         &  & \multicolumn{1}{l}{\textbf{89.8}}  & \multicolumn{1}{l}{\textbf{89.0}}  & \multicolumn{1}{l}{\textbf{89.4}} & \multicolumn{1}{l}{} & \multicolumn{1}{l}{\textbf{79.0}}     & \multicolumn{1}{l}{\textbf{78.0}}     & \multicolumn{1}{l}{\textbf{78.5}} \\ \bottomrule
\end{tabular}}}
\caption{Model comparisons on CoNLL04. $\vartriangle$ denotes using the micro-averaged F1. $\blacktriangle$ denotes using the macro-averaged F1. Bold values denote the state-of-the-art {results}.}\label{table2}
\end{table}

\subsection{Main results}
Table \ref{table1}, Table \ref{table2}, and Table \ref{table3} show the model comparison results. We have the following observations: 
(1) Our best model consistently surpasses all the selected baselines in terms of F1. 
(2) On ACE05, compared to the strongest baselines \cite{chendanqi,renli}, our best model obtains +0.7\%, +1.9\%, and +1.1\% F1 gains on NER, RE, and RE+, resepectively.
(3) On CoNLL04, compared to the strongest baselines \cite{zhao,cabot}, our best model obtains +1.0\% and +1.7\% \textbf{micro-averaged} F1 gains on NER and RE+, respectively. And compared to the strongest baselines \cite{eb_ul,wa_lu}, our model obtains +2.5\% and +3.1\% \textbf{macro-averaged} F1 gains on NER and RE, respectively.
(4) On ADE (\textbf{without overlapping entities}), compared to the strongest baseline \cite{yanzhi}, our best model obtains +1.0\% and +1.3\% F1 gains on NER and RE, respectively.
(5) On the full ADE (\textbf{with overlapping entities}), compared to the strongest baseline \cite{lai}, our best model obtains +1.1\% and +2.7\% F1 gains on NER and RE, respectively.

We attribute these performance gains to: (1) The success of using token-level label information in span-based joint extraction. (2) The bi-directional information interaction between NER and RE. (3) The effectiveness of the extended BIO tagging scheme.
Additionally, we report concrete positive and negative case studies to help understand our model, as shown in Section \ref{4.5}.

\begin{table}[h]
\centering
\setlength{\tabcolsep}{1.5mm}{
\begin{tabular}{lrlccccccc}
\toprule
\multirow{2}{*}{Model}                                                                             & \multirow{2}{*}{PLM}    &  & \multicolumn{3}{c}{NER}                                                  &                      & \multicolumn{3}{c}{RE+}                                                        \\ \cmidrule{4-6} \cmidrule{8-10} 
                                                                                                   &                         &  & P                     & R                     & F1                       &                      & P                        & R                        & F1                       \\ \midrule
Eberts \textit{et al.} \cite{eb_ul} $\spadesuit $ \ \ \ \ & \texttt{BERT}                    &  & 89.0                  & 89.6                  & 89.3                     &                      & 77.8                     & 78.0                     & 78.8                     \\
Ji \textit{et al.} \cite{ji} $\spadesuit$                        & \texttt{BERT}                    &  & 89.9                  & 91.3                  & 90.6                     &                      & 79.6                     & 81.9                     & 80.7                     \\
Lai \textit{et al.} \cite{lai} $\spadesuit$                      & \texttt{BERT}                    &  & -                     & {-}            & 90.7                     &                      & -                        & {-}               & 81.7                     \\
Li \textit{et al.} \cite{li16}                                   & -                       &  & 79.5                  & 79.6                  & 79.5                     &                      & 64.0                     & 62.9                     & 63.4                     \\
Li \textit{et al.} \cite{li17}                                   & -                       &  & 82.7                  & 86.7                  & 84.6                     &                      & 67.5                     & 75.8                     & 71.4                     \\
Bekoulis \textit{et al.} \cite{be}                               & -                       &  & 84.7                  & 88.2                  & 86.4                     &                      & 72.1                     & 77.2                     & 74.6                     \\
Eberts \textit{et al.} \cite{eb_ul}                             & \texttt{BERT}                    &  & 89.3                  & 89.3                  & 89.3                     &                      & 78.1                     & 80.4                     & 79.2                     \\
Zhao \textit{et al.} \cite{zhao}                                 & \texttt{ELMo}                    &  & -                     & -                     & 89.4                     &                      & -                        & -                        & 81.1                     \\
Yan \cite{yanzhi}                                                                 & \texttt{BERT}                    &  & {-}            & {-}            & 89.6                     &                      & -                        & {-}               & 80.0                     \\
Wang and Lu \cite{wa_lu}                                                         & \texttt{ALBERT}                  &  & -                     & {-}            & 89.7                     &                      & -                        & {-}               & 80.1                     \\
Shen \textit{et al.} \cite{shen}                                 & \texttt{BERT}                    &  & 89.5                  & 91.3                  & 90.4                     &                      & 84.2                     & 83.4                     & 80.7                     \\
Cabot \textit{et al.} \cite{cabot}                                                                & \texttt{BART}                    &  & {-} & {-} & {-}    & \multicolumn{1}{l}{} & {81.5} & {83.1} & {82.2} \\
Yan \cite{yanzhi}                                                                 & \texttt{ALBERT}                  &  & {-}            & {-}            & 91.3                     &                      & -                        & {-}               & 83.2                     \\ \midrule
STSN (Ours)                                                                                          & {\texttt{BERT}}   &  & 91.3                      & 91.9                      & 91.6                     &                      &82.8                          &\textbf{84.6}                          & 83.7                     \\
STSN (Ours)                                                                                          & {\texttt{ALBERT}} &  &\textbf{91.5}                       &\textbf{93.1}                       & \textbf{92.3}                     &                      &\textbf{84.8}                          &  84.2                        & \textbf{84.5}                     \\  \midrule
STSN (Ours)  $\spadesuit$                                                                                        & {\texttt{BERT}}                        &  &90.8                       &  91.4                     & 91.1                     &                      &83.3                          & 83.7                         & 83.5                     \\
STSN (Ours)   $\spadesuit$                                                                                       &  {\texttt{ALBERT}}                       &  & \multicolumn{1}{l}{\textbf{91.6}}  & \multicolumn{1}{l}{\textbf{92.0}}  & \multicolumn{1}{l}{\textbf{91.8}} & \multicolumn{1}{l}{} & \multicolumn{1}{l}{\textbf{85.0}}     & \multicolumn{1}{l}{\textbf{83.8}}     & \multicolumn{1}{l}{\textbf{84.4}} \\ \bottomrule
\end{tabular}}
\caption{Model comparisons on ADE. ${\spadesuit}$ denotes evaluating STSN on the full ADE dataset (with overlapping entities). Bold values denote the state-of-the-art results.}\label{table3}
\end{table}

\subsection{Analysis}

We report analysis results on the dev set of ACE05 and the test sets of CoNLL04 and ADE.\footnote{Following previous work \cite{eb_ul,ji,wa_lu,zhao}, we combine the training and dev sets of CoNLL04 to train our STSN. Thus we use the test set for the analysis. And since ADE does not contain a dev set, we also use the test set for the analysis.}
And we take SpERT \cite{eb_ul} as the baseline, which is the closest span-based model to ours. 
SpERT uses BERT as the encoder and two linear decoders to classify spans and span pairs.  
For a fair comparison, we use our BERT-based STSN.

\subsubsection{Performance against entity length}
\begin{figure}[htb]
\centering
\includegraphics[width=0.82\textwidth]{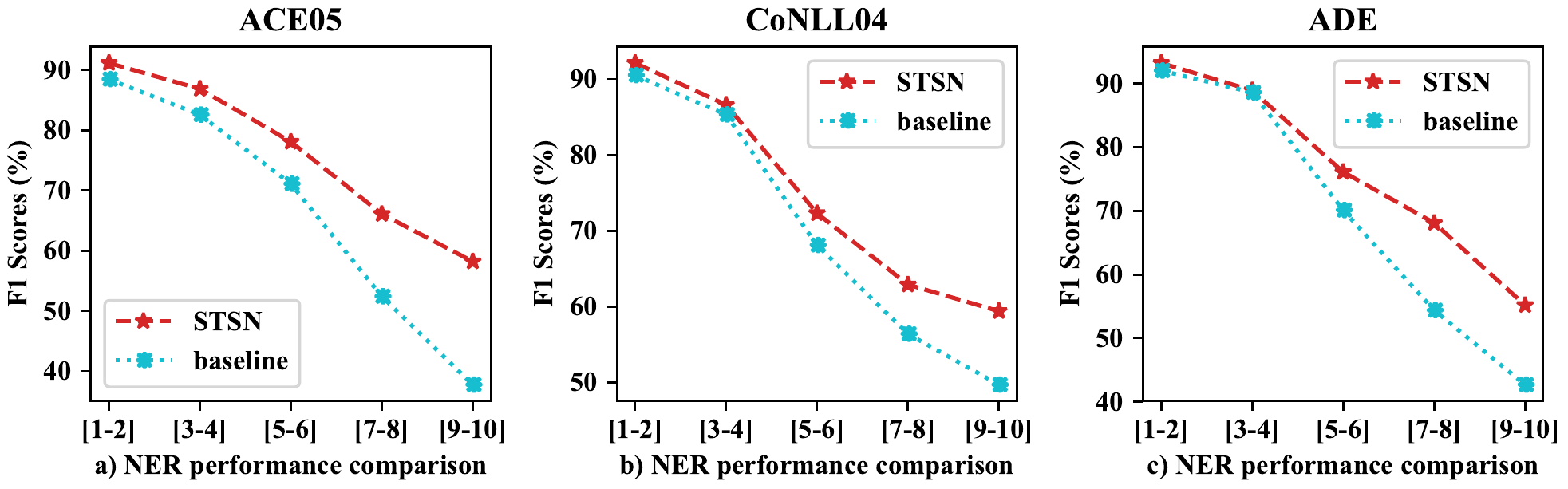}
\caption{NER performance comparisons under various grouped entity lengths across the three datasets. }
\label{figure5}
\end{figure}

Figure \ref{figure5} shows performance comparisons on NER under various entity lengths. We divide all entity lengths, which is restricted by span width threshold $\epsilon$, into [1-2], [3-4], [5-6], [7-8], and [9-10]. We have the following observations: across all datasets, (1) STSN consistently outperforms the baseline under all length intervals. (2) Performance improvements brought by STSN are generally further enhanced when the entity length increases. Specifically, STSN obtains much higher F1 gains under [7-8] and [9-10] than the ones under [1-2] and [3-4], demonstrating that STSN is more effective in terms of long entities.

\subsubsection{Performance against text length}
\begin{figure}[H]
\centering
\includegraphics[width=0.82\textwidth]{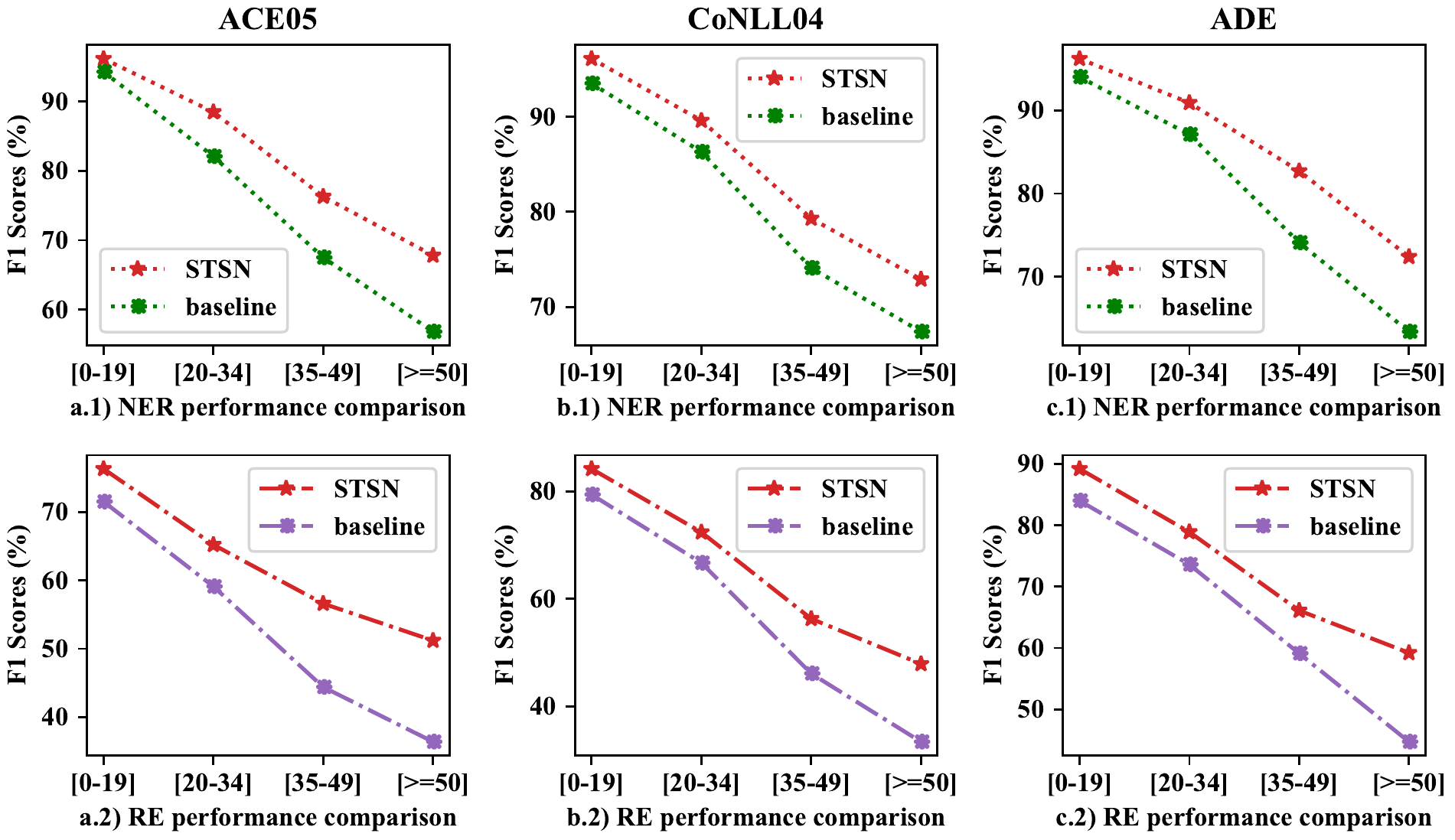}
\caption{NER and RE performance comparisons under various grouped text lengths across the three datasets.}\label{figure6}
\end{figure}

We compare STSN with the baseline under grouped text lengths. As Figure \ref{figure6} shows, we divide text lengths into [0-19], [20-34], [35-49], and [$\geq$50]. We have the following observations: across the three datasets, (1) STSN performs way better than the baseline under all text lengths on both NER (a.1, b.1, and c.1) and RE (a.2, b.2, and c.2). (2) Performance gains brought by STSN are generally further enhanced when text length increases. In particular, STSN obtains the best performance gains under [$>$=50] on both NER and RE, demonstrating that STSN is more effective in terms of long texts. 

\subsection{Ablation study}
We conduct ablation studies on our BERT-based STSN and report ablation results on the dev set of ACE05 and the test sets of CoNLL04 and ADE.

\subsubsection{Ablations on various attention layers} We conduct ablations on attention layer numbers by deep stacking various attention layers in STSN. Table \ref{table4} shows the ablation results, from which we can observe that: across the three datasets, (1) When deep stacking three attention layers, STSN performs the best (82.2\% Ave. F1). (2) STSN with only one attention layer performs the worst, which we attribute to the fact that one layer cannot fully infuse token-level label information into token semantic representations. (3) When the number of attention layers increases, the model performance generally first drastically increases then slightly decreases. We attribute this to the fact that deeper models make it easier to fully infuse token-level label information into token semantic representations, while much deeper models tend to infuse more noisy information, which harms the model performance.

\begin{table}[h]
\renewcommand{\arraystretch}{0.78}
\centering
\begin{tabular}{lccccccc}
\toprule
\multirow{2.3}{*}{Method} & \multicolumn{2}{c}{ACE05}                                                                                                                       & \multicolumn{2}{c}{CoNLL04}                                                                                                                     & \multicolumn{2}{c}{ADE}                                                                                                                         & \multicolumn{1}{c}{\multirow{2.3}{*}{Ave. F1}} \\ \cmidrule(lr){2-3} \cmidrule(lr){4-5} \cmidrule(lr){6-7}
                        & \multicolumn{1}{c}{\begin{tabular}[c]{@{}c@{}}NER\end{tabular}} & \multicolumn{1}{c}{\begin{tabular}[c]{@{}c@{}}RE+\end{tabular}} & \multicolumn{1}{c}{\begin{tabular}[c]{@{}c@{}}NER\end{tabular}} & \multicolumn{1}{c}{\begin{tabular}[c]{@{}c@{}}RE+\end{tabular}} & \multicolumn{1}{c}{\begin{tabular}[c]{@{}c@{}}NER\end{tabular}} & \multicolumn{1}{c}{\begin{tabular}[c]{@{}c@{}}RE+\end{tabular}} & \multicolumn{1}{c}{}                         \\ \midrule
STSN + deep stacking    &                                                                        &                                                                        &                                                                        &                                                                        &                                                                        &                                                                        &                                              \\
\ \ \texttt{1 AttentionLayer}       & 87.6                                                                   & 59.2                                                                   & 87.4                                                                   & 72.1                                                                   & 88.9                                                                   & 81.1                                                                   &79.4                                              \\
\ \ \texttt{2 AttentionLayers}       & 88.7                                                                   & 60.5                                                                   & 90.0                                                                   & 73.9                                                                   & 89.5                                                                   & 80.4                                                                   &80.5                                              \\
\ \ \texttt{3 AttentionLayers}       & 89.5                                                                   & 62.6                                                                   & 90.9                                                                   & 75.0                                                                   & 91.6                                                                   & 83.7                                                                   & \textbf{82.2}                                         \\
\ \ \texttt{4 AttentionLayers}       & 89.2                                                                   & 62.5                                                                   & 91.3                                                                   & 75.2                                                                   & 90.5                                                                   & 83.8                                                                   &82.1                                              \\
\ \ \texttt{5 AttentionLayers}       & 88.9                                                                   & 62.6                                                                   & 90.4                                                                   & 74.2                                                                   & 90.7                                                                   & 83.2                                                                   &81.7                                              \\
\ \ \texttt{6 AttentionLayers}      & 89.1                                                                   & 62.0                                                                   & 90.4                                                                   & 74.4                                                                   & 90.5                                                                   & 82.9                                                                   &81.6                                              \\ \bottomrule
\end{tabular}
\caption{Ablation study on attention layer numbers. We solely report the F1 scores and consider the averaged score of the 6 F1 scores in each row to be \textbf{Ave. F1}, which we use as an overall evaluation metric. The bold value denotes the best result.}\label{table4}
\end{table}

\subsubsection{Ablations on model components}\label{4.4.2}
\begin{figure}[]
	\centering
	\begin{minipage}[t]{0.325\linewidth}
		\centering
		\includegraphics[width=2in]{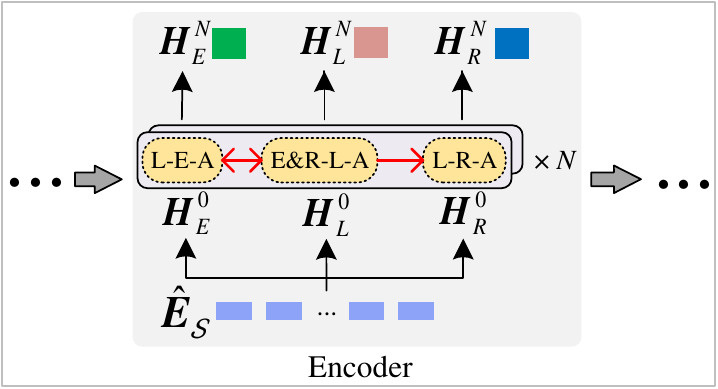}
		\caption{Removing the information flow from RE to NER, as the red lines show.}
		\label{figure7}
	\end{minipage}
	\begin{minipage}[t]{0.327\linewidth}
		\centering
		\includegraphics[width=2in]{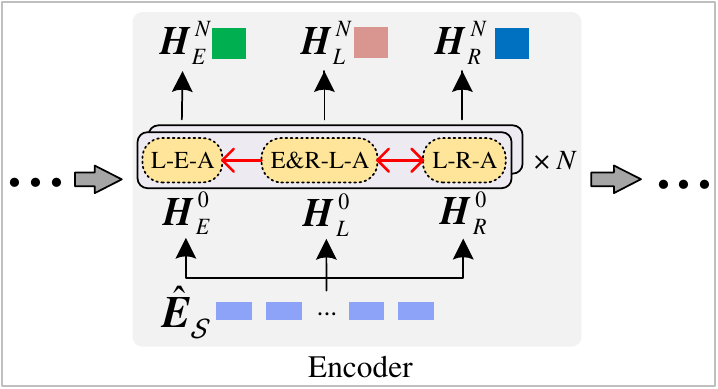}
		\caption{Removing the information flow from NER to RE, as the red lines show.}
		\label{figure8}
	\end{minipage}
	\begin{minipage}[t]{0.32\linewidth}
		\centering
		\includegraphics[width=2in]{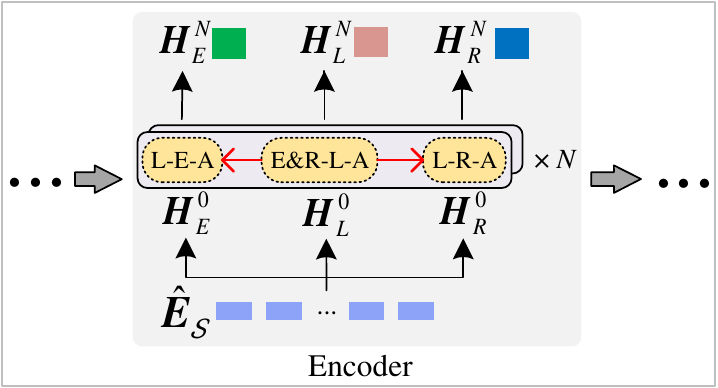}
		\caption{Removing the information interactions, as the red lines show.}
		\label{figure9}
	\end{minipage}
\end{figure}

Table \ref{table5} reports the ablation results across the three datasets, where 
\begin{itemize}
\item[(1)] {``\texttt{w/o Label}'' denotes ablating token-level label information. We realize this ablation by removing the stacked attention layers and the decoder for sequence tagging-based NER from STSN. After doing this, our model cannot benefit from the token-level label information. The results show that using the token-level label information boosts the model performance by delivering +2.7\% to +3.1\% F1 gains on NER and +4.2\% to +6.0\% F1 gains on RE+.}
\item[(2)] {``\texttt{w/o Bi\mbox{-}Interaction}$\dag$'' denotes removing the information flow from RE to NER but keeping the information flow from NER to RE, as shown in Figure \ref{figure7}. We realize this ablation by making the $\mathbi{K}$ and $\mathbi{V}$ of E\&R-L-A be $\mathbi{H}_E$ and the $\mathbi{Q}$ of E\&R-L-A be $\mathbi{H}_L$. Thus $\mathbi{H}_L$ no longer attends to $\mathbi{H}_R$ and solely attends to $\mathbi{H}_E$. The results show that the information flow from RE to NER brings +1.1\% and +0.8\% averaged F1 gains on NER and RE, respectively.}
\item[(3)] {``\texttt{w/o Bi\mbox{-}Interaction}$\ddag$'' denotes removing the information flow from NER to RE but keeping the information flow from RE to NER, as shown in Figure \ref{figure8}. We realize this ablation by making the $\mathbi{K}$ and $\mathbi{V}$ of E\&R-L-A be $\mathbi{H}_R$ and the $\mathbi{Q}$ of E\&R-L-A be $\mathbi{H}_L$. Thus $\mathbi{H}_L$ no longer attends to $\mathbi{H}_E$ and solely attends to $\mathbi{H}_R$. The results show that the information flow from NER to RE brings +0.4\% and +1.3\% averaged F1 gains on NER and RE, respectively. }
\item[(4)] {``\texttt{w/o Interaction}'' denotes removing the information interactions between NER and RE, as shown in Figure \ref{figure9}. We realize this ablation by making the $\mathbi{Q}$, $\mathbi{K}$, and $\mathbi{V}$ of E\&R-L-A be $\mathbi{H}_L$. In other words, E\&R-L-A is the self-attention in the current scenario, disabling the information interactions between NER and RE. The results show that the bi-directional information interactions bring +1.2\% and +1.5\% averaged F1 gains on NER and RE, respectively.}
\end{itemize}

Based on these observations, we can conclude that the performance gains mainly benefit from using the token-level label information, revealing that our motivation is sufficient. Moreover, the bi-directional information interaction is consistently superior to the two unidirectional information flows, validating the effectiveness of our novel bi-directional design.

\begin{table}[H]
\renewcommand{\arraystretch}{0.78}
\centering
\begin{tabular}{lllllll}
\toprule
\multirow{2.3}{*}{Method} & \multicolumn{2}{c}{ACE05}                                                                                                                       & \multicolumn{2}{c}{CoNLL04}                                                                                                                     & \multicolumn{2}{c}{ADE}                                                                                                                         \\ \cmidrule(lr){2-3} \cmidrule(lr){4-5} \cmidrule(lr){6-7}
                        & \multicolumn{1}{c}{\begin{tabular}[c]{@{}c@{}}NER\end{tabular}} & \multicolumn{1}{c}{\begin{tabular}[c]{@{}c@{}}RE+\end{tabular}} & \multicolumn{1}{c}{\begin{tabular}[c]{@{}c@{}}NER\end{tabular}} & \multicolumn{1}{c}{\begin{tabular}[c]{@{}c@{}}RE+\end{tabular}} & \multicolumn{1}{c}{\begin{tabular}[c]{@{}c@{}}NER\end{tabular}} & \multicolumn{1}{c}{\begin{tabular}[c]{@{}c@{}}RE+\end{tabular}} \\ \midrule
STSN                    & 89.5                                                                   & 62.6                                                                   & 90.9                                                                   & 75.0                                                                   & 91.6                                                                   & 83.7                                                                   \\
\ \ \texttt{w/o Label}               & 86.4 (-3.1)                                                                   & 56.6 (-6.0)                                                                   & 88.1 (-2.7)                                                                 & 70.8 (-4.2)                                                                   & 88.7 (-2.9)                                                                   & 78.5 (-5.2)                                                                   \\
\ \ \texttt{w/o Bi-Interaction}$\dag$      & 89.0 (-0.5)                                                                   & 61.6 (-1.0)                                                                   & 89.6 (-1.3)                                                                   & 74.4 (-0.6)                                                                   & 90.1 (-1.5)                                                                   & 82.9 (-0.8)                                                                   \\ 
\ \ \texttt{w/o Bi-Interaction}$\ddag$      & 89.2 (-0.3)                                                                   & 61.4 (-1.2)                                                                   & 90.2 (-0.7)                                                                   & 73.5 (-1.5)                                                                   & 91.5 (-0.1)                                                                   & 82.4 (-1.3)                                                                   \\ 
\ \ \texttt{w/o Interaction}      & 88.9 (-0.6)                                                                   & 61.7 (-0.9)                                                                   & 89.5 (-1.4)                                                                   & 73.4 (-1.6)                                                                   & 90.1 (-1.5)                                                                   & 81.6 (-2.1)                                                                   \\ 
\bottomrule
\end{tabular}
\caption{Ablation results. We solely report the F1 scores. The values in parentheses denote the F1 score decreases (compared to STSN) caused by corresponding ablation settings.}\label{table5}
\end{table}

\subsection{Case study}\label{4.5}
We conduct qualitative analysis on concrete examples to help understand our model. We take SpERT as the baseline, which is the closest span-based model to ours. For a fair comparison, we use our BERT-based STSN.

\subsubsection{Positive example} 
Table \ref{table6} reports four positive examples. 
In Text 1, SpERT mistakenly predicts ``House of Delegates'' as a \texttt{LOC} entity, while STSN correctly predicts it as a \texttt{ORG} entity. We attribute it to the fact that STSN enables span representations to incorporate token-level label information in the case that STSN correctly tags ``House of Delegates'' with \texttt{ORG} labels. Moreover, STSN correctly predicts that $<$``House of Delegates'', ``Maryland''$>$ holds a \texttt{OrgBased\_In} relation. Text 2 shows a similar example, where STSN correctly predicts ``La.'' as a \texttt{LOC} entity, as well as the \texttt{Located\_In} relation hold by $<$``Grand Isle'', ``La.''$>$.

Text 3 and 4 mainly show the effects of using token-level label information in relation classification. For example, both SpERT and STSN correctly predict all entities of Text 3, but SpERT mistakenly predicts that there is no relation between these entities. In contrast, STSN correctly predicts the two \texttt{Located\_In} relations. We attribute it to using token-level label information in relation representations, enabling our model to know detailed entity types beforehand.

\begin{table*}[h]
\centering
\resizebox{1\linewidth}{!}{
\begin{tabular}{l|l|l}
 \toprule 
\multicolumn{2}{c|}{\textbf{Text 1}}         & \textcolor[RGB]{0,0,255}{Judith Toth} says she returned for a fourth term in \textcolor[RGB]{0,0,255}{Maryland}'s \textcolor[RGB]{0,0,255}{House of Delegates}                                          \\ \midrule
\multirow{2.3}{*}{{SpERT}} & Entity       & {[}{Judith Toth}]$_{\texttt{PER}}$\quad  {[}{Maryland}]$_{\texttt{LOC}}$\quad  {[}{House of Delegates}]$_{{\textcolor[RGB]{255,0,0}{\texttt{LOC}}}}$ \\    \cmidrule{2-3} 
                                & Relation     & \textless{}{House of Delegates}, {Maryland}, $\textcolor[RGB]{255,0,0}{\texttt{Located\_In}}$\textgreater{}                                                         \\ \midrule
\multirow{3.5}{*}{{STSN}}  & Token label   & \texttt{B-PER\ \ I-PER\ \ O\ \ O\ \ O\ \ O\ \ O\ \ O\ \ O\ \ O\ \ B-LOC\ \ B-ORG\ \ I-ORG\ \ I-ORG} \\ \cmidrule{2-3} 
                                & Entity       & {[}{Judith Toth}]$_{\texttt{PER}}$\quad  {[}{Maryland}]$_{\texttt{LOC}}$ \quad  {[}{House of Delegates}]$_{\texttt{ORG}}$                                                     \\ \cmidrule{2-3} 
                                & Relation     & \textless{}{House of Delegates}, {Maryland}, $\texttt{OrgBased\_In}$\textgreater{}                                                        \\ \midrule
\multicolumn{2}{c|}{\textbf{Text 2}}         & One man was lost from an oil rig off \textcolor[RGB]{0,0,255}{Grand Isle}, \textcolor[RGB]{0,0,255}{La.}, as the storm moved in \\ \midrule
\multirow{2.3}{*}{{SpERT}} & Entity       & {[}{Grand Isle}]$_{\texttt{LOC}}$
\\    \cmidrule{2-3} 
                                & Relation     & \textcolor[RGB]{255,0,0}{{{No relation}}}                                                                                                                                                                               \\ \midrule
\multirow{3.5}{*}{{STSN}}  & Token label   & \texttt{O\ \ O\ \ O\ \ O\ \ O\ \ O\ \ O\ \ O\ \ O\ \ B-LOC\ \ I-LOC\ \ B-LOC\ \ O\ \ O\ \ O\ \ O\ \ O     }                                                                     \\ \cmidrule{2-3} 
                                & Entity       & {[}{Grand Isle}]$_{\texttt{LOC}}$\quad  {[}{La.}]$_{\texttt{LOC}}$ %
 \\ \cmidrule{2-3} 
                                & Relation     & \textless{}{Grand Isle}, {La.}, $\texttt{Located\_In}$\textgreater{}                                                        \\ \midrule

\multicolumn{2}{c|}{\textbf{Text 3}}         & \textcolor[RGB]{0,0,255}{Seattle} has a hour-glass figure, squeezed between \textcolor[RGB]{0,0,255}{Puget Sound} and \textcolor[RGB]{0,0,255}{Lake Washington}                                         \\ \midrule
\multirow{2.3}{*}{{SpERT}} & Entity       & {[}Seattle{]}$_{\texttt{LOC}}$\quad  {[}Puget Sound{]}$_{\texttt{LOC}}$\quad  {[}Lake Washington{]}$_{\texttt{LOC}}$                                                           \\ \cmidrule{2-3} 
                                & Relation     &    \textcolor[RGB]{255,0,0}{{{No relation}}}                                                                                                                       \\ \midrule
\multirow{3.5}{*}{{STSN}}  & Token label & \texttt{B-LOC\ \ O\ \ O\ \ O\ \ O\ \  O\ \ O\ \ B-LOC\ \ I-LOC\ \ O\ \ B-LOC\ \ I-LOC}                                                                           \\ \cmidrule{2-3} 
                                & Entity       & {[}Seattle{]}$_{\texttt{LOC}}$ \quad{[}Puget Sound{]}$_{\texttt{LOC}}$ \quad{[}Lake Washington{]}$_{\texttt{LOC}}$                                                           \\ \cmidrule{2-3} 
                                & Relation     & \textless{}Puget Sound, Seattle, $\texttt{Located\_In}$\textgreater{}, \textless{}Lake Washington, Seattle, $\texttt{Located\_In}$\textgreater{} \\ \midrule 

\multicolumn{2}{c|}{\textbf{Text 4}}         &An enraged \textcolor[RGB]{0,0,255}{Khrushchev} instructed \textcolor[RGB]{0,0,255}{Soviet} ships to ignore \textcolor[RGB]{0,0,255}{Kennedy}'s naval blockade \\ \midrule
\multirow{2.3}{*}{{SpERT}} & Entity       & {[}{Khrushchev}]$_{\texttt{PER}}$\quad  {[}{Soviet}]$_{\texttt{LOC}}$ \quad  {[}{Kennedy}]$_{{{\texttt{PER}}}}$ 
\\    \cmidrule{2-3} 
                                & Relation    & \textcolor[RGB]{255,0,0}{{{No relation}}}                                                                                                                                                                               \\ \midrule
\multirow{3.5}{*}{{STSN}}  & Token label   &  \texttt{O\ \ O\ \ B-PER\ \ O\ \ B-LOC\ \ O\ \ O\ \ O\ \ B-PER\ \ O\ \ O }                                                                     \\ \cmidrule{2-3} 
                                & Entity       & {[}{Khrushchev}]$_{\texttt{PER}}$\quad  {[}{Soviet}]$_{\texttt{LOC}}$ \quad  {[}{Kennedy}]$_{{{\texttt{PER}}}}$ \\ \cmidrule{2-3} 
                                & Relation     & \textless{}{Khrushchev}, {Soviet}, $\texttt{Live\_In}$\textgreater{}                                                        \\ \bottomrule
\end{tabular}}
\caption{Positive examples regarding using token-level label information in the span-based joint extraction, where the \textcolor{red}{red font} denotes that entities or relations are mistakenly predicted, and the \textcolor[RGB]{0,0,255}{blue font} denotes corresponding entities located in texts, and all labels, entities, and relations in the STSN rows are predicted correctly.}\label{table6}
\end{table*}

\subsubsection{Negative example} 
{We also report a negative example, as Table \ref{table7} shows. In this example, STSN mistakenly tags a token label: ``president'' is tagged with \texttt{I-ORG}, which is supposedly tagged with \texttt{O}. However, STSN still correctly predicts all entities and relations of this Text. Moreover, we find that STSN successfully tackles most of the similar cases (97.56\%) across the three datasets. We attribute it to the fact that STSN learns only to incorporate useful label information, enabling our model to avoid suffering from wrong label predictions.}

\begin{table*}[h]
\centering
\resizebox{1\linewidth}{!}{
\begin{tabular}{l|l|l}
 \toprule 
\multicolumn{2}{c|}{\textbf{Text}}         & But \textcolor[RGB]{0,0,255}{Jack Frazier}, \textcolor[RGB]{0,0,255}{Rotary Club} president, said volunteers picked up the ducks                                          \\ \midrule
\multirow{2.3}{*}{{SpERT}} & Entity       & {[}{Jack Frazier}]$_{\texttt{PER}}$\quad  {[}{Rotary Club}]$_{\texttt{ORG}}$ \\    \cmidrule{2-3} 
                                & Relation     & \textless{}{Jack Frazier}, {Rotary Club}, {\texttt{Work\_For}}$\textgreater{}$                                                         \\ \midrule
\multirow{3.5}{*}{{STSN}}  & Token label   & \texttt{O\ \ B-PER\ \ I-PER\ \ B-ORG\ \ I-ORG}\ \ \texttt{\textcolor{red}{I-ORG}}\ \ \texttt{O\ \ O\ \ O\ \ O\ \ O\ \ O}                                                                     \\ \cmidrule{2-3} 
                                & Entity       & {[}{Jack Frazier}]$_{\texttt{PER}}$\quad  {[}{Rotary Club}]$_{\texttt{ORG}}$                                                       \\ \cmidrule{2-3} 
                                & Relation     & \textless{}{Jack Frazier}, {Rotary Club}, {\texttt{Work\_For}}$\textgreater{}$                                                         \\ \bottomrule 
\end{tabular}}
\caption{A negative example, in which {STSN} mistakenly predicts a token label (i.e., the \texttt{\textcolor{red}{I-ORG}}), but it still correctly predicts all entities and relations. 
}\label{table7}
\end{table*}

\section{Conclusion}
In this paper, we propose a Sequence Tagging augmented Span-based Network (STSN) for the joint entity and relation extraction task. 
STSN enables the span-based joint extraction model to use token-level label information, which is achieved by deep stacking multiple attention layers. Moreover, STSN establishes bi-directional information interactions between NER and RE, which is proven effective.
Furthermore, we extend the BIO tagging scheme, allowing STSN to use the label information of overlapping entities. 
Experiments on three datasets show that STSN consistently outperforms other competing models in terms of F1.
Since STSN only considers the two-fold overlapping entities, we will investigate upgrading our model in the future to extract other overlapping entities.

\Acknowledgements{This work was supported by Hunan Provincial Natural Science Foundation (Grant Nos. 2022JJ30668 and 2022JJ30046). }


\begin{thebibliography}{99}

\bibitem{li_ji} Li Q, Ji H. Incremental joint extraction of entity mentions and relations. In: Proceedings of the 52nd Annual Meeting of the Association for Computational Linguistics, Baltimore, Maryland, 2014. 402–412

\bibitem{mi_ba}  Miwa M, Bansal M. End-to-end relation extraction using LSTMs on sequences and tree structures. In: Proceedings of the 54th Annual Meeting of the Association for Computational Linguistics, Berlin, Germany, 2016. 1105–1116

\bibitem{ka_ca} Katiyar A, Cardie C. Going out on a limb: Joint extraction of entity mentions and relations without dependency trees. In: Proceedings of the 55th Annual Meeting of the Association for Computational Linguistics, Vancouver, Canada, 2017. 917–928

\bibitem{wei} Ye W, Li B, Xie R, et al. Exploiting entity BIO tag embeddings and multi-task learning for relation extraction with imbalanced data. In: Proceedings of the 57th Annual Meeting of the Association for Computational Linguistics, Florence, Italy, 2019. 1351–1360

\bibitem{lin} Lin Y, Ji H, Huang F, et al. A joint neural model for information extraction with global features. In: Proceedings of the 58th Annual Meeting of the Association for Computational Linguistics, Online, 2020. 7999–8009

\bibitem{luan18} Luan Y, He L, Ostendorf M, et al. Multi-task identification of entities, relations, and coreference for scientific knowledge graph construction. In: Proceedings of the 2018 Conference on Empirical Methods in Natural Language Processing, Brussels, Belgium, 2018. 3219–3232 

\bibitem{di_al} Dixit K, Al-Onaizan Y. Span-level model for relation extraction. In: Proceedings of the 57th Annual Meeting of the Association for Computational Linguistics, Florence, Italy, 2019. 5308–5314

\bibitem{eb_ul} Eberts M, Ulges A. Span-based joint entity and relation extraction with transformer pre-training. In: Proceedings of the 24th European Conference on Artificial Intelligence, Santiago de Compostela, Spain, 2020. 1–8

\bibitem{ji} Ji B, Yu J, Li S, et al. Span-based joint entity and relation extraction with attention-based span-specific and contextual semantic representations. In: Proceedings of the 28th International Conference on Computational Linguistics, Online, 2020. 88–99

\bibitem{chendanqi} Zhong Z, Chen D. A frustratingly easy approach for entity and relation extraction. In: Proceedings of the 2021 Conference of the North American Chapter of the Association for Computational Linguistics: Human Language Technologies, Online, 2021. 50–61

\bibitem{luan} Luan Y, Wadden D, He L, et al. A general framework for information extraction using dynamic span graphs. In: Proceedings of the 2019 Conference of the North American Chapter of the Association for Computational Linguistics: Human Language Technologies, Minneapolis, Minnesota, 2019. 3036–3046 

\bibitem{wadden} Wadden D, Wennberg U, Luan Y, et al. Entity, relation, and event extraction with contextualized span representations. In: Proceedings of the 2019 Conference on Empirical Methods in Natural Language Processing and the 9th International Joint Conference on Natural Language Processing, Hong Kong, China, 2019. 5784–5789

\bibitem{docred} Yao Y, Ye D, Li P, et al. DocRED: A large-scale document-level relation extraction dataset. In: Proceedings of the 57th Annual Meeting of the Association for Computational Linguistics, Florence, Italy, 2019. 764–777 

\bibitem{tacred} Zhang Y, Zhong V, Chen D, et al. Position-aware attention and supervised data improve slot filling. In: Proceedings of the 2017 Conference on Empirical Methods in Natural Language Processing, Copenhagen, Denmark, 2017. 35–45

\bibitem{nyt} Riedel S, Yao L, McCallum A. Modeling relations and their mentions without labeled text. In: Proceedings of the Joint European Conference on Machine Learning and Knowledge Discovery in Databases, Berlin, Heidelberg, 2010. 148–163

\bibitem{webnlg} Zeng X, Zeng D, He S, et al. Extracting relational facts by an end-to-end neural model with copy mechanism. In: Proceedings of the 56th Annual Meeting of the Association for Computational Linguistics, Melbourne, Australia, 2018. 506–514

\bibitem{semeval} Hendrickx I, Kim S, Kozareva Z, et al. SemEval-2010 task 8: Multi-way classification of semantic relations between pairs of nominals. In: Proceedings of the 5th International Workshop on Semantic Evaluation, Uppsala, Sweden, 2010. 33–38

\bibitem{conll} Roth D, Yih W. A linear programming formulation for global inference in natural language tasks. In: Proceedings of the 8th Conference on Computational Natural Language Learning at HLT-NAACL, Boston, Massachusetts, USA, 2004. 1–8

\bibitem{ade} Gurulingappa H, Rajput A, Roberts A. Development of a benchmark corpus to support the automatic extraction of drug-related adverse effects from medical case reports. J Biomed Inform, 2012, 45: 885–892

\bibitem{be} Bekoulis G, Deleu J, Demeester T, et al. Joint entity recognition and relation extraction as a multi-head selection problem. Expert Syst Appl, 2018, 114: 34-45.

\bibitem{zhao} Zhao S, Hu M, Cai Z, et al. Modeling dense cross-modal interactions for joint entity-relation extraction. In: Proceedings of the 29th International Joint Conference on Artificial Intelligence, Online, 2020. 4032–4038

\bibitem{lee1} Lee K, He L, Zettlemoyer L. Higher-order coreference resolution with coarse-to-fine inference. In: Proceedings of the 2018 Conference of the North American Chapter of the Association for Computational Linguistics: Human Language Technologies, New Orleans, Louisiana, 2018. 687–692 

\bibitem{he} He L, Lee K, Levy O, et al. Jointly predicting predicates and arguments in neural semantic role labeling. In: Proceedings of the 56th Annual Meeting of the Association for Computational Linguistics, Melbourne, Australia, 2018. 364–369

\bibitem{elmo} Peters M, Neumann M, Iyyer M, et al. Deep contextualized word representations. In: Proceedings of the 2018 Conference of the North American Chapter of the Association for Computational Linguistics: Human Language Technologies, New Orleans, Louisiana, 2018. 2227–2237

\bibitem{bert} Devlin J, Chang M, Lee K, et al. BERT: pre-training of deep bidirectional transformers for language understanding. In: Proceedings of the 2019 Conference of the North American Chapter of the Association for Computational Linguistics: Human Language Technologies, Minneapolis, Minnesota, 2019. 4171–4186 

\bibitem{albert} Lan Z, Chen M, Goodman S, et al. ALBERT: a lite BERT for self-supervised learning of language representations. In: Proceedings of the 8th International Conference on Learning Representations, Online, 2020. 1-17

\bibitem{wordpiece} Wu Y, Schuster M, Chen Z, et al. Google’s neural machine translation system: bridging the gap between human and machine translation. 2016. ArXiv:1609.08144

\bibitem{transformer} Vaswani A, Shazeer N, Parmar N, et al. Attention is all you need. In: Proceedings of the 31st Conference on Neural Information Processing Systems, Long Beach, CA, USA, 2017. 5998–6008

\bibitem{ace} Doddington G, Mitchell A, Przybocki M, et al. The automatic content extraction (ACE) program – tasks, data, and evaluation. In: Proceedings of the 4th International Conference on Language Resources and Evaluation, Lisbon, Portugal, 2004. 837-840

\bibitem{wa_lu} Wang J, Lu W. Two are better than one: Joint entity and relation extraction with table-sequence encoders. In: Proceedings of the 2020 Conference on Empirical Methods in Natural Language Processing, Online, 2020. 1706–1721 

\bibitem{sun} Sun C, Wu Y, Lan M, et al. Extracting entities and relations with joint minimum risk training. In: Proceedings of the 2018 Conference on Empirical Methods in Natural Language Processing, Brussels, Belgium, 2018. 2256–2265

\bibitem{li} Li X, Yin F, Sun Z, et al. Entity-relation extraction as multi-turn question answering. In: Proceedings of the 57th Annual Meeting of the Association for Computational Linguistics, Florence, Italy, 2019. 1340–1350

\bibitem{shen} Shen Y, Ma X, Tang Y, et al. A trigger-sense memory flow framework for joint entity and relation extraction. In: Proceedings of the Web Conference 2021, Online, 2021. 1704-1715

\bibitem{renli} Ren L, Sun C, Ji H, et al. HySPA: hybrid span generation for scalable text-to-graph extraction. In: Findings of the Association for Computational Linguistics: ACL-IJCNLP, Online, 2021. 4066–4078

\bibitem{ng_ve} Nguyen D Q, Verspoor K. End-to-end neural relation extraction using deep biaffine attention. In: Proceedings of the 41st European Conference on Information Retrieval, Cologne, Germany, 2019. 729–738

\bibitem{mi_sa} Miwa M, Sasaki Y. Modeling joint entity and relation extraction with table representation. In: Proceedings of the 2014 Conference on Empirical Methods in Natural Language Processing, Doha, Qatar, 2014. 1858–1869

\bibitem{zhang} Zhang M, Zhang Y, Fu G. End-to-end neural relation extraction with global optimization. In: Proceedings of the 2017 Conference on Empirical Methods in Natural Language Processing, Copenhagen, Denmark, 2017. 1730–1740 

\bibitem{cabot} Huguet P, Navigli R. REBEL: relation extraction by end-to-end language generation. In: Findings of the 2021 Conference on Empirical Methods in Natural Language Processing, Punta Cana, Dominican Republic, 2021. 2370–2381

\bibitem{yanzhi} Yan Z, Zhang C, Fu J, et al. A partition filter network for joint entity and relation extraction. In: Proceedings of the 2021 Conference on Empirical Methods in Natural Language Processing, Punta Cana, Dominican Republic, 2021. 185–197

\bibitem{lai} Lai T, Ji H, Zhai C, et al. Joint biomedical entity and relation extraction with knowledge-enhanced collective inference. In: Proceedings of the 59th Annual Meeting of the Association for Computational Linguistics and the 11th International Joint Conference on Natural Language Processing, Online, 2021. 6248–6260

\bibitem{li16} Li F, Zhang Y, Zhang M, et al. Joint models for extracting adverse drug events from biomedical text. In: Proceedings of the 25th International Joint Conference on Artificial Intelligence, New York, USA, 2016. 2838–2844

\bibitem{li17} Li F, Zhang M, Fu G, et al. A neural joint model for entity and relation extraction from biomedical text. BMC bioinformatics, 2017, 18: 1–11

\end{thebibliography}

\end{document}